# Autonomous Driving–5 Years after the Urban Challenge: The Anticipatory Vehicle as a Cyber-Physical System

Christian Berger[1] and Bernhard Rumpe[2]

[1]Department of Computer Science and Engineering
Chalmers | University of Gothenburg
Sweden
christian.berger@chalmers.se

[2]Software Engineering
RWTH Aachen University
Germany
rumpe@se-rwth.de

**Abstract:** In November 2007 the international competition DARPA Urban Challenge took place on the former George Airforce Base in Victorville, California to significantly promote the research and development on autonomously driving vehicles for urban environments. In the final race only eleven out of initially 89 competitors participated and "Boss" from Carnegie Mellon University succeeded. This paper summarizes results of the research carried out by all finalists within the last five years after the competition and provides an outlook where further investigation especially for software engineering is now necessary to achieve the goal of driving safely and reliably through urban environments with an anticipatory vehicle for the mass-market.

## 1 Introduction

Autonomous driving was significantly fostered within that last decade due to three major challenges, which were carried out by DARPA–the research agency of the Department of Defense. The most exciting one was the 2007 DARPA Urban Challenge, in which competitors from all over the world had to develop robotic vehicles that were able to drive entirely without any human interaction in urban environments. Compared to the previous Grand Challenges from 2004 and 2005, autonomously driving vehicles had not only to deal with other moving vehicles but they also had to obey the Californian traffic law during all their autonomous operations.

In Fig. 1(a) the autonomously driving vehicle "Caroline" [RBL[+]08] developed from the Technische Universität Braunschweig under the organization of the authors is depicted. For that vehicle, the problem to drive without any human interaction was split into three main tasks according to Fig. 1(b): Data perception and preprocessing from all sensors to generate an environmental model, understanding this surroundings' model to derive the next deriving decision, and performing actions within the system's context i.e. steering and accelerating the vehicle. The results gathered during that competition meanwhile led to the development of her successor "Leonie" [NHO[+]11].

Current driverless vehicles have collected data from more than 140,000mi [Thr10a, Thr10b] or are already running in China [Nan11]. Although that competition was named "Urban

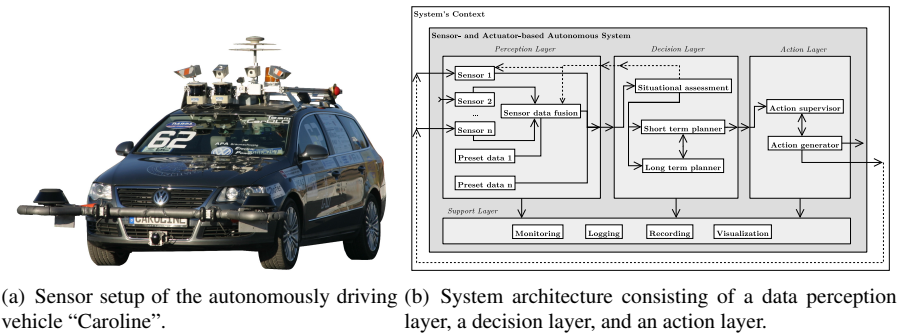

(a) Sensor setup of the autonomously driving vehicle "Caroline".

(b) System architecture consisting of a data perception layer, a decision layer, and an action layer.

Figure 1: "Caroline"–the contribution from Technische Universität Braunschweig competing in the 2007 DARPA Urban Challenge and its principal system architecture.

Challenge" pedestrians and bicyclists had not been regarded at all. Thus, what has been achieved within the last five years after that spectacular final event and where are still aspects, which need more attention to further reduce traffic jams, to save fuel, and most important to prevent casualties and fatalities? This contribution summarizes research results from finalist teams from the last five years with a strong focus on aspects that were not explicitly included in the DARPA Urban Challenge competition.

## 2 Localization & Perceiving the Vehicle's Surroundings

As shown in Fig. 1(a), a set of redundant sensors with overlapping viewing areas is necessary to create a reliable representation from the autonomously driving vehicle's surroundings. Besides this sensors' online data, preprocessed offline data improves the vehicle performance. [SU09] outlines an aerial image analysis, which relies on self-supervised machine learning to detect parking spots. From this analysis the fundamental topological structure for the area of interest is derived to provide an augmented graph as drivable paths in advance.

Due to limited accuracy or temporal lack of GPS, the precise position of the autonomously driving vehicle is fundamentally important. [LT10] as an extension to [LMT07] outlines an approach that models the static environment as a probabilistic grid with a $0.15 \text{x} 0.15 \text{m}^2$ resolution where every cell represents a Gaussian distribution over infra-red remittance. Thus during online localization, an RMS-accuracy of nearly 0.1m could be achieved. However, the outlined implementation requires approximately 10MB/mi of storage and is not entirely independent from rough weather conditions or modifications to the environment (e.g. construction sites); moreover, elevation information could improve this map.

During the competition neither traffic lights nor traffic signs had to be detected. Nowadays, static traffic signs are provided as annotations within digital maps and the state of traffic lights can be provided wirelessly by vehicle-to-X. However for the latter, the infras-tructure itself must be modified, which would be very expensive (e.g. nearly 2,200 traffic lights exist in Berlin); an optical detection would cover already existing traffic lights. Thus, [LADT11] outlines a map-based approach to control the region of interest for detecting the position and orientation of a traffic light within camera images; [FU11] additionally encodes semantics (i.e. various flashing states) or labels like "dim" to indicate traffic lights that are difficult to detect in images to parametrize an algorithm. However, these approaches also unveil problems due to lens flares or heavy rain.

As depicted in Fig. 1(a), a static sensor set is mounted and thus, viewing angles are fixed and could not be adapted where necessary. In [SU08], a pointable sensor for covering unobserved areas of interest is outlined, which is adjusted by an information entropy model of an intersection's uncertainty for example. This approach could be extended to model temporary occluded areas by e.g. dynamic objects. An alternative approach is outlined by [SP12] that proposes the combination of sensors with vehicle-to-X-communication.

The competition excluded explicitly pedestrians and bicyclists in general; however, for the broader usage of driverless vehicles these at least protected road users[1] must be detected reliably. Contrary to detecting pedestrians continuously within the driving area, [BCG$^+$09] presents an approach that focuses on specific situations like stopped vehicles or crosswalks detected by a laser scanner. Within these pre-detected areas, pedestrians are validated by the computationally intense vision system. Contrary to pedestrians, bicyclists may also share the road with vehicles; in [CRZ11] a monocular vision-based approach is described, which uses a deformable shape model for detection and an interacting multiple model filter to track a bicyclist. In [WGR11b], a real-time detecting approach based on contour cues is outlined running at 20fps on VGA resolution and which could be improved by using a GPU. However, vision-based approaches depend on external illuminating and would be insufficient to operate reliably.

From a software engineering point of view, we will face the challenge that the mass-market will have sensors in many different forms and qualities. Furthermore, degradation of their quality as well as replacement by other sensors with different quality characteristics will be immanent. This poses special adaptivity to the software architecture that processes sensor data, both on easy adaptation and on self-awareness of the quality characteristics of the sensors in the car. Graceful degradation of software functions based on sudden failures of sensors need also to be handled reliably.

## 3  Understanding the Surroundings' Data

The sensors perceive a vehicle's surroundings, which must be analyzed and interpreted in real-time. In [HW11], a layer on top of the perception layer is introduced that abstracts details, which are currently irrelevant to the situation analysis and maps objects to a graph-based environmental model. Thus, deriving decisions is reduced to the objects, which are relevant to future actions like turning while yielding right of way to forthcoming vehicles. In [CWW11], an environmental graph-structure is also outlined that extended the road graph used during the competition. Their representation includes annotations about way point interpolations, lane categories, and bridges or tunnels to be suitable for German autobahn and in urban environments [WGR11a].

---

[1] According to Statistisches Bundesamt, 4,002 fatally injured road users (398 bicyclists) were counted in 2011.

As outlined in Sec. 2, aerial images could be preprocessed to provide maps for parking lots. In [SUWL10] these maps are used to generate proper drivable paths from an unstructured parking area. In contrast, [DTMD09] uses a Voronoi field to model path lengths and distances to obstacles for generating drivable trajectories. A more elaborated approach is described in [JHMBH10] that could also be applied when GPS is not available as in parking garages.

Reliable detection of pedestrians and bicyclists is crucial during the decision process. [ZRG+09] outlines an approach, which predicts their possible future trajectories. A similar focus is provided by [GSLS11], in which plausible future motions for each tracked person are calculated. A more generic approach is outlined by [HC11] that uses hierarchical trajectory clustering to improve the computational effort. The goal is to cluster obstacle trajectories, which have an effect on the planned vehicle's trajectory.

[VGZD07] describes an approach for high-level situation reasoning that tries to predict the future evolvement of a situation. The main idea is to assess the current traffic situation for finding a similar previous case to derive a proper decision. Hereby, a case is defined by the behavior of the autonomous vehicle, the behavior of other participants, and an estimation how similar the current traffic situation to that case is.

In [WDSL11], a Markov Decision Process is used to derive single-lane driving decisions; thus, uncertainties for a sensor's noise and the other vehicles' behavior can be modeled, which results in a robust driving behavior in uncertain situations compared to prior ACC systems. A more elaborated system for assisting the driver on highways is presented in [SFG+10]. This system evaluates continuously the vehicle's surroundings to warn the driver about unsafe lane change situations or to predict safe lane change speeds.

Recent research efforts also include data exchange realized by vehicle-to-X communication. [NHF+11] therefore proposes an approach that explicitly includes this data into a so called knowledge layer for deriving driving decisions. The main idea is to combine sensor-gathered information with wirelessly received information to generate a more reliable representation of the vehicle's surroundings. Thus, information from occluded situations non-visible to sensors can be considered.

From a software engineering point of view, the adaptivity on the sensor side will enforce high-adaptivity on the sensor data fusion and understanding. This enforces a reliable and flexible software architecture that processes data on several levels, but also understands to shortcut processing in case of emergencies (e.g. when suddenly braking is necessary). While the general architecture from Fig. 1(b) has been used in many projects, these architectures are generally not flexible enough and usually do not provide data connection shortcuts. Furthermore, many of today's experimental architectures for such an anticipatorily driving vehicle assume a rather centralized form of processing while today's car manufacturers' architectures use massively distributed control unit structures. Today's vehicle software architecture is still very much based on functions and therefore does not enable reuse of software components. Software architecture needs to become independent of function architecture, decoupled by adequate and high-level software interfaces and decomposed in fine grained, reusable building blocks of software.

## 4 Acting within the Vehicle's Context

After evaluating the vehicle's surroundings the derived driving decision needs to be carried out. In [WZKT10] a semi-reactive approach for generating trajectories is outlined. Its input are abstract commands from the previous layer, which are used to calculate a desired trajectory with respect to long-term goals like keeping a desired velocity while following other vehicles, and to short-term goals like avoiding collisions.

In [LF09] an approach is outlined to generate dynamically feasible trajectories with the goal to travel on high speed. It is based on a lattice state space running with real-time performance by using a combination of high-resolution action space around the vehicle and a low-resolution action space elsewhere. In contrast, [ZS09] presents a method that reduces the required amount of nodes to model all possible vehicle's motions. An approach to stabilize state trajectories for inner-city speeds up to $6\frac{m}{s}$ is outlined in [WGB10].

In [KPJ$^+$10], a system is described, which allows to control precisely an autonomously driving vehicle to slide into a sideways parking spot. Although there is only a limited practical usage for the daily usage in urban environments, the authors showed that they were able to control the vehicle repeatedly even in this extremely dynamical maneuver.

From a software engineering point of view, modeling of control algorithms is very mature. Challenges are the reliability even in degrading situations, in particular as we cannot reliably predict the absent driver (e.g. due cognitive distraction) to take over within seconds as a fallback. Low-level safety and reliability is the key here. As a second challenge, we face that adaptivity of higher functions to customer specific needs enforce capabilities of dynamic updates and enhancements of car software. Among others this imposes an additional security problem. Thus, we need an appropriate software architecture that allows to aggregate and cummulate information from real and virtual sensors, but also produces efficient shortcuts from sensors to actuators in case of emergencies.

## 5 Monitoring & Evaluating the Vehicle's Performance

As shown in Fig. 1(b) the development of autonomously driving vehicles is also supported by visualizing and analyzing recorded data from various test drives to comprehend a driving decision. However, this data can only be analyzed when all required sensors are mounted and calibrated accordingly. But for testing the first sketches of algorithms anyhow, existing data sets from the DARPA Urban Challenge [HAO$^+$10] or from Ford [PME11] are available to validate the correctness of data handling and processing. Furthermore, the Stanford Track Collection [TLT11] also provides nearly 14,000 labeled tracks for further inspection. These tracks are recorded using a Velodyne HDL-64E S2 LIDAR for various street scenarios. For example, this database could be used to develop algorithms that were able to classify traffic participants like pedestrians and bicyclists to evaluate the reliability of another detecting sensor like a vision system.

From a software engineering point of view, this availability of data provides an interesting opportunity to define an enhanced process to develop software. New versions of software can be tested in virtual environments completely detached from hardware. This enables software to drive many millions virtual miles much faster than any "real time" HiL could allow. Simulation is the key for software quality management and software and function architectures can be defined in such a way that various combination of software components and their interplay can be tested while its context is being simulated through appropriate mocks.

## 6  Engineering the Autonomously Driving Vehicle's Software

The previous sections summarized the research results accounting aspects, which were beyond the requirements of the DARPA Urban Challenge according to the general system architecture. However from a software engineer's point of view, aspects like the development process itself, integration and quality assurance for components or the entire system, or the re-use of existing components come into mind. However, these aspects have hardly emerged from finalist teams within the last five years. Only the authors of [BDWL11] address the actual reuse of software components that is related to software engineering.

However, concerning the increasing complexity and uncertainty of sensed input data, real test drives as carried out to find optimal parameters for the 2007 competition are insufficient. Instead, a thorough engineering approach to develop the software of these systems is required, which supports the development already at early stages by virtual environments. Regarding the tight schedule for the 2007 DARPA Urban Challenge together with the limited vehicles' surroundings, the main focus of all teams was on the actual performance during the qualification and the final. However, modern comfort and active safety systems has to operate properly even in hardly foreseeable traffic situations. Thus, today's software engineering must be extended by approaches, which explicitly embrace simulation environments especially for sensor-based data.

In [Ber10] and [BR12], foundations, modeling aspects, and applications for such a virtual environment are outlined and discussed. One of the main ideas is to model the vehicle's surroundings in terms of a single point of truth. From this hierarchical environmental model, particular aspects for the relevant data processing layers (cf. Fig. 1(b)) are derived and used within a dedicated simulation context: Its object-oriented representation is used to serve the decision layer, while the annotated various 3D models are used to generate the required raw sensor data for the perception layer. During the development of "Caroline" and a succeeding project at the University of California, Berkeley, selected aspects from this simulation-based software engineering approach were evaluated.

# 7 Conclusion and Outlook

In this paper, contributions from finalist teams of the 2007 DARPA Urban Challenge from the last five years were summarized with a focus on aspects, which were explicitly excluded from that competition. Regarding the goal to develop an anticipatorily driving vehicle that is reliable enough for the daily use, current research is on a solid way. This is also documented by an increasing number of advanced driver assistance and safety systems, which increase the safety not only for the occupants but also for other traffic participants like pedestrians. In the upcoming future, we will encounter more and more systems that base on a reliable environmental model and which will be increasingly interconnected to serve their users.

These assisting and interconnected sensor- and actor-based systems are currently classified as *Cyber-Physical Systems*, which enable newly arising services and business models to alleviate our daily life [GB12]. However, the engineering of these increasingly intelligent, interconnected, and autonomously acting systems demands for more elaborated and partially new techniques and methods [GRSS12]; also formal methods can be successfully applied to focus on the *correct* implementation of the *right* requirements [SHGB11].

For the long-term success of these systems, it is fundamentally important to not only regard the desired behavior and functionality of these systems but even more to analyze and model *timing properties* as part of their semantic correctness within their real domain [Lee12]. Without these adapted or newly developed methods it will hardly be possible to evaluate the correct behavior of a system within its application domain. Furthermore in future, the evaluation of a system in reality must be extended by virtual approaches to ensure a valid behavior in critical or unforeseeable situations. This trend can already be seen by the research's results in the evolution of autonomously driving vehicles.